\documentclass{article}
\usepackage{spconf,amsmath,graphicx}

\usepackage{comment}
\usepackage{multirow} % table
\usepackage{color}
\usepackage{amssymb} % \checkmark, mathbb
\newcommand{\bvec}[1]{\mbox{\boldmath $#1$}}

\newcommand{\ts}{\textsuperscript} % \ts{10} -> 10^th
\newcommand{\bhline}[1]{\noalign{\hrule height #1}}
\usepackage{arydshln}            % Draw dotted line in table
\usepackage[subrefformat=parens]{subcaption} % Subcaption
\usepackage{hyperref} % e-mail

\usepackage{fancyhdr} % Prepare "fancyhdr.sty"
\renewcommand{\headrulewidth}{0pt}
\renewcommand\footrule{\hrule height 0.5pt \vspace{2.5mm}\hrule height 0pt}
\title{HOKEM: Human and Object Keypoint-based Extension Module\\for Human-object Interaction Detection}

\name{Yoshiki Ito}
\address{R\&D Group, Hitachi, Ltd., Japan\\
\normalsize{\texttt{yoshiki.ito.xf@hitachi.com}}}
\begin{document}

\ninept
\maketitle

%\setlength{\abovedisplayskip}{1pt} % top margin
%\setlength{\belowdisplayskip}{1pt} % bottom margin

%#####################################################################################################

% 100 to 150 words
% Be identical to the abstract text submitted electronically along with the paper cover sheet.
% No indent !

\begin{abstract}

\noindent
Human-object interaction (HOI) detection for capturing relationships between humans and objects is an important task in the semantic understanding of images.
When processing human and object keypoints extracted from an image using a graph convolutional network (GCN) to detect HOI, it is crucial to extract appropriate object keypoints regardless of the object type and to design a GCN that accurately captures the spatial relationships between keypoints.
This paper presents the human and object keypoint-based extension module (HOKEM) as an easy-to-use extension module to improve the accuracy of the conventional detection models.
The proposed object keypoint extraction method is simple yet accurately represents the shapes of various objects.
Moreover, the proposed human-object adaptive GCN (HO-AGCN), which introduces adaptive graph optimization and attention mechanism, accurately captures the spatial relationships between keypoints.
Experiments using the HOI dataset, V-COCO, showed that HOKEM boosted the accuracy of an appearance-based model by a large margin.

\end{abstract}


%#####################################################################################################

%\vspace{3mm}
\section{INTRODUCTION}
\label{s:intro}

Advances in neural networks have led to rapid progress in computer vision tasks such as object detection~\cite{liu2022convnet} and human action recognition~\cite{kong2022human}.
For a better understanding of images, it is necessary to focus not only on each person or object, but also on the relationships between them.
Thus, the detection of human-object interaction (HOI) has been highlighted as one of the important tasks in the advanced semantic understanding of images.

HOI detection is a task that aims to localize a human-object pair and clarify their relationship as a triplet $<$\textit{human-relation-object}$>$.
Appearance-based methods have been proposed to identify the position of both a person and an object and to estimate their interaction~\cite{gkioxari2018detecting,gao2018ican,kim2020uniondet,ulutan2020vsgnet}.
Since accurate human pose estimation has become feasible~\cite{cao2017realtime,geng2021bottom}, several methods have been proposed to extract human pose-based features and utilize them in conjunction with appearance-based features~\cite{zhou2019relation,li2019transferable,li2020pastanet,wan2019pose,zhong2020polysemy,kim2020detecting}.
Furthermore, some methods also extract object keypoints in addition to human keypoints~\cite{liang2020pose,liu2022human,zhu2022skeleton}.
In the skeleton-based action recognition task~\cite{yan2018spatial,shi2019two,song2021constructing}, pose features are processed by a graph convolutional network (GCN) assuming the structure of the human body as a graph.
Subsequently, in~\cite{liang2020pose,zhu2022skeleton}, human and object keypoints have been processed by GCN, resulting in more accurate HOI detection.

When using object keypoints for HOI detection, it is crucial to extract keypoints that sufficiently represent the features of the object.
In~\cite{liang2020pose}, only the center of bounding box was selected as an object keypoint.
To represent the size and shape of objects, object keypoint exraction methods have been proposed in~\cite{liu2022human,zhu2022skeleton}.
In~\cite{liu2022human}, each bounding box is divided into $3 \times 3$ grids and one keypoint is extracted from each grid on the basis of instance segmentation.
Whereas, by applying a skeletonization algorithm~\cite{zhang1984fast} to the whole region obtained from instance segmentation, the method of extracting nine keypoints has been proposed~\cite{zhu2022skeleton}.
However, not all important keypoints of an object may be contained in each square of the grid that divides the box evenly.
In addition, 
%because a wide variety of objects are targeted, 
it is difficult for a pose estimation method, such as that used for humans, to fully represent the shapes of diverse objects.

The design of GCN is also important for highly accurate HOI detection when processing human and object keypoints with GCN.
In the above mentioned studies~\cite{liang2020pose,zhu2022skeleton}, because keypoints were processed by shallow GCNs with one and three layers respectively, it is not sufficient for capturing the relationships between human and object.
Moreover, a mechanism to optimize the topology of the predefined graph should be introduced because GCN requires a predefined graph to represent the connection between keypoints but there is no real connection between the human body and objects.
Furthermore, the attention mechanism should be introduced into the network to obtain keypoints to focus on.

To address the above points, this paper presents a human and object keypoint-based extension module (HOKEM) as an easy-to-use extension module to improve the accuracy of the conventional appearance-based HOI detection models.
The module contains two main proposals: the object keypoint extraction method and GCN.
The contributions of this paper are summarized as follows.
\begin{itemize}
	\item HOKEM can be easily utilized with a variety of conventional appearance-based HOI detection models to obtain higher accuracy as it simply multiplies the output probabilities.
	\item A novel object keypoint extraction method accurately represents the shapes of a wide variety of objects.
	\item A GCN, which introduces adaptive graph optimization and an attention mechanism, accurately extracts the relationships between human and object keypoints.
	\item Experiments using the HOI dataset, V-COCO, showed that HOKEM significantly boosted the accuracy of an appearance-based detection model by approximately 5.0 mAP.
\end{itemize}

%#####################################################################################################

\begin{figure*}[t!]
\centering
	%\vspace{-1mm}
	\includegraphics[width=0.9\linewidth]{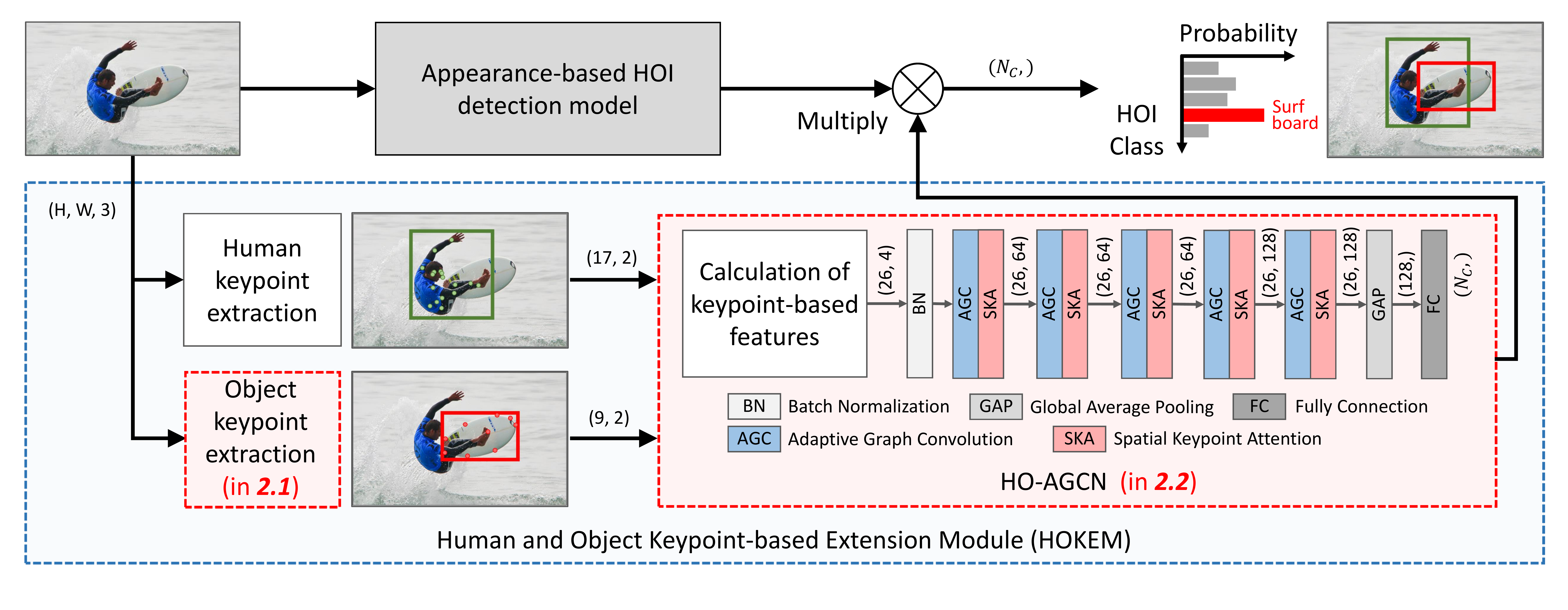}
	\vspace{-6mm}
	\caption{Overall architecture of HOKEM and how it is used with the appearance-based HOI detection model. $H, W, N_c$ denote the height and width of input image, and the number of HOI classes, respectively. There are 17 human keypoints and 9 object keypoints in this figure.}
	\label{f:overall}
	\vspace{-3mm}
\end{figure*}

\vspace{-2mm}
\section{Proposed Method}
\label{s:method}
\vspace{-2mm}

The overall architecture of the proposed extension module, HOKEM, and how it is used with the conventional HOI detection model are shown in Figure~\ref{f:overall}.
A significant advantage of HOKEM is that it can be easily utilized with a wide variety of models to obtain higher accuracy
because the final classification probability is calculated as the product of the probabilities output by the conventional detection model and the proposed extension module.
%
%Because the final classification probability
% for a detection task 
%is calculated as the product of the probabilities output by the conventional detection model and HOKEM, a significant advantage of HOKEM is that it can be easily utilized with a wide variety of models to obtain higher accuracy.
%
HOKEM consists of
(1) a novel object keypoint extraction method using instance segmentation
and
(2) a human-object adaptive graph convolutional network (HO-AGCN) for accurate HOI detection, which will be described in~\ref{ss:okp} and~\ref{ss:ho-agcn}, respectively.

\vspace{-3mm}
\subsection{Object Keypoint Extraction}
\label{ss:okp}
\vspace{-2mm}

In order for the GCN to process a human body and an object, appropriate keypoints are needed to represent the shapes of both.
There is a wide variety of object types, and it is not possible to define object keypoints as those which are meaningful for humans, e.g., elbow, knee.
Thus, this paper proposes an object keypoint extraction method to accurately represent object shapes regardless of their type.

The object keypoints extracted by the proposed method are shown in Figure~\ref{f:test2} (c).
First, object regions are extracted from an image by instance segmentation as shown in (b).
Next, the center of gravity, the top, bottom, left, and right most coordinates of the region are determined as keypoints (green and blue points in (c)).
In order to more accurately represent the object contour, the intermediate coordinates of two adjacent outer circumference points, e.g., top and left, are calculated.
Then, a linear function connecting gravity and the intermediate coordinates can be calculated.
A new keypoint is selected as the coordinate closest to the box that is determined to be the object area on the linear function line.
This is performed for all pairs of two adjacent points to obtain four new keypoints (yellow points in (c)).
To make a set of object keypoints, the above nine points are arranged in order such as $V_o = [\rm{gravity}, \rm{top}, \rm{left}, \cdots]$.
By clarifying the location of the center of gravity and precisely representing the contour of the object, an accurate spatial representation of the object is possible, regardless of the type of object.

\vspace{-3mm}
\subsection{HO-AGCN}
\label{ss:ho-agcn}
\vspace{-1mm}

%\vspace{-3mm}
\subsubsection{Human-object Graph}
\label{sss:ho-graph}
\vspace{-2mm}

A spatial graph $G=(V,E)$ is constructed from human and object keypoints for graph convolution.
$V$ is represented as a set of human and object keypoints $V_h$ and $V_o$, where $V_h = \{v_i|i=1,2,\cdots,N_h\}$ and $V_o = \{v_i|i=1,2,\cdots,N_o\}$.
$N_h$ and $N_o$ denote the number of human and object keypoints, respectively.
In this method, $N_o$ is set as 9 as described in~\ref{ss:okp}.

\begin{figure}[t!]
\centering
	%\vspace{-1mm}
	\includegraphics[width=0.9\linewidth]{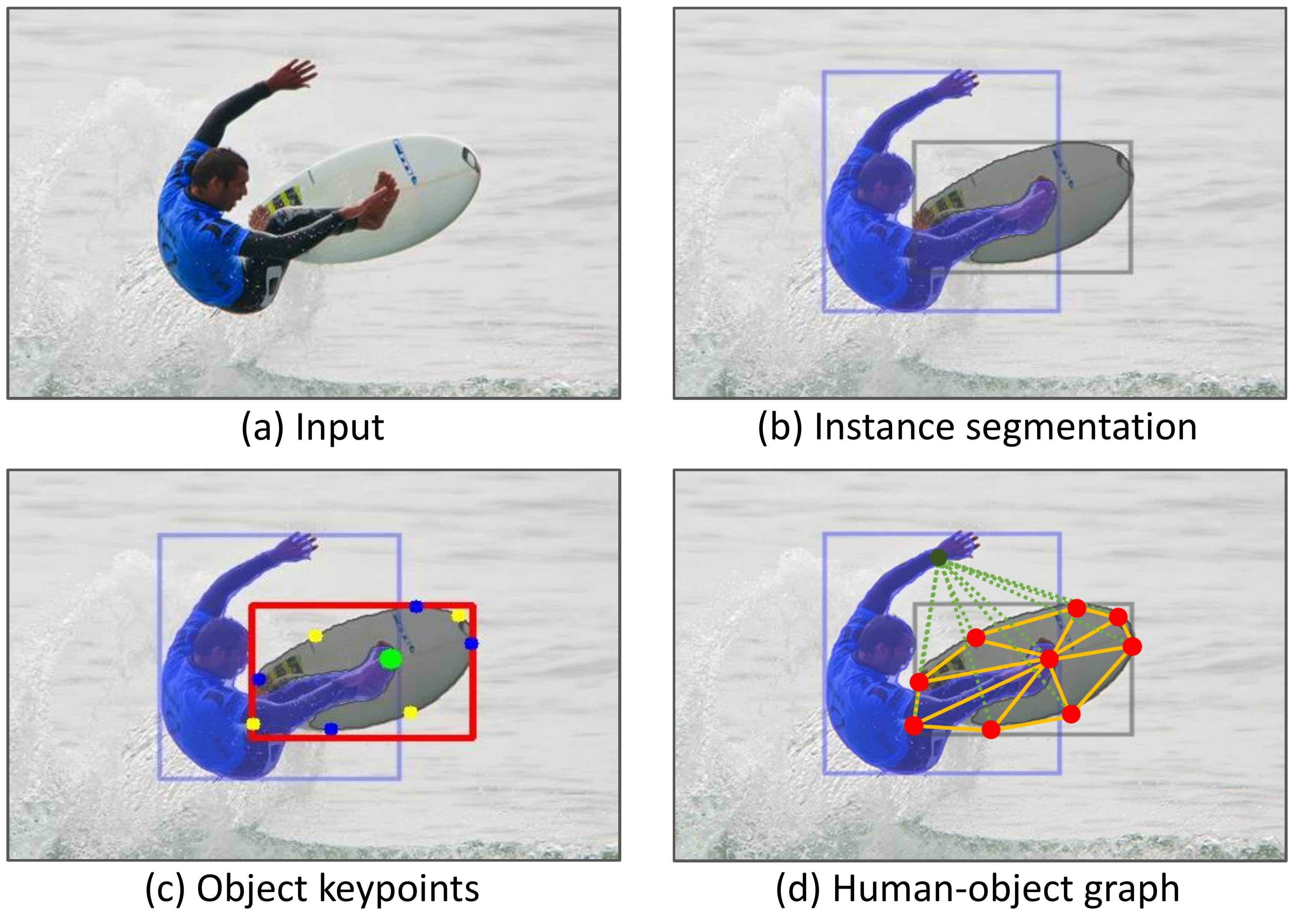}
	\vspace{-2mm}
	\caption{Object keypoints extracted by the proposed method and human-object graph. In (d), only edges between object keypoints and edges from one human keypoint to object keypoints are illustrated for visibility.}
	\label{f:test2}
	\vspace{-5mm}
\end{figure}

The proposed human-object graph is constructed as shown in Figure~\ref{f:test2} (d).
$E$ is a set of edges that are connected from three perspectives: human, object, and human-object.
In the human edges, connections are made based on the physical structure of the human body using $V_h$, as in the conventional skeleton-based action recognition method~\cite{yan2018spatial,shi2019two,song2021constructing}.
In the object edges, keypoints around the object periphery from $V_o$ are first connected.
Next, all of the periphery points are connected to the gravity of the object.
In the human-object edges, all of the object keypoints connect to six representative joints in the human body (head, torso, and distal portion of the extremities).
The human keypoints are connected to the object keypoints, taking into account the structure of the human body and object, resulting in highly accurate graph convolution.

\vspace{-3mm}
\subsubsection{Keypoint-based Features}
\label{sss:input}
\vspace{-2mm}

The proposed method first normalizes the bounding boxes by the size of the upper body for robust HOI detection against differences in the height and depth of a person from the camera.
All elements of $V$ are subtracted on the basis of the upper left coordinate of the human bounding box and then divided by the diagonal length of the box surrounded by the shoulders and hips.

Next, distance and angle from the neck to each keypoint are calculated from $V$.
Given $C = \{c_i = (x_i, y_i)|i=1,2,\cdots,N_v\}$, where $N_v=N_h+N_o$, as a set of human and object keypoint coordinates,
the proposed method calculates the distance $d_i$ and angle $a_i$ for a keypoint $v_i$ as
\begin{align}%eqnarray}
\label{e:feat}
	\begin{aligned}
		d_i &= \sqrt{(x_i - x_n)^2 + (y_i - y_n)^2}, \\
		a_i &= \arctan \left| \frac{y_i - y_n}{x_i - x_n} \right| \times \frac{2}{\pi},
	\end{aligned}
\end{align}%eqnarray}
where $c_n = (x_n, y_n)$ denotes the coordinate of the person's neck.
Finally, the input feature of a keypoint $v_i$ is concatenated as $\bvec{f}_\textrm{in}(v_i) = [x_i, y_i, d_i, a_i] \in \mathbb{R}^4$ and input into the network.
%
%Finally, the feature for a keypoint $v_i$ input into the first later is concatenated as $\bvec{f}_\textrm{in}(v_i)^{(1)} = [c_i, d_i, a_i] \in \mathbb{R}^4$.

%a_i &= \left| \tan^{-1} \frac{y_i - y_n}{x_i - x_n} \right| \times \frac{2}{\pi}.
% https://www.monster-dive.com/blog/web_creative/20161219_000293.php
% https://daeudaeu.com/c-atan/

\vspace{-3mm}
\subsubsection{Graph Convolution and Attention Mechanism}
\label{sss:conv}
\vspace{-2mm}

The principal layers utilized in HO-AGCN are adaptive graph convolution (AGC) layer and spatial keypoint attention (SKA) layer as shown in Figure~\ref{f:overall}.
Given $\mathcal{S}_i$ as a set of adjacent keypoints for the $i$-th keypoint $v_i$ and an input feature $\bvec{f}_\textrm{in}(v_j)$ from a keypoint $v_j$ in $\mathcal{S}_i$, the graph convolution operation on keypoint $v_i$ is written as
\begin{align}
\label{e:gc}
	\bvec{f}_\textrm{out}(v_i) = \sum_{v_j \in \mathcal{S}_i} \frac{1}{Z_{ij}} \
	\bvec{f}_\textrm{in}(v_j) \cdot w(l_i(v_j)).
\end{align}
A spatial partitioning strategy for distinguishing node characteristics is set in the same manner as that of spatial configuration partitioning~\cite{yan2018spatial}; that is, the convolution kernel size is set to 3 and $\mathcal{S}_i$ is divided into three groups according to the distance between the node and the center of gravity.
$l_i(v_j)$ denotes the label map at keypoint $v_j$ and is determined by the partitioning strategy.
$w(\cdot)$ denotes the function for obtaining the weight on the basis of the kernel index.
The normalizing term $Z_{ij}$ denotes the cardinality of the subsets to balance the contributions of different subsets.

The proposed method utilizes AGC~\cite{shi2019two} to optimize the topology of the predefined graphs and lower the dependence on the graphs.
\mbox{Eq. (\ref{e:gc})} is transformed as
\begin{align}%eqnarray}
\label{e:agc}
	\bvec{F}_\textrm{out} = \sum_{k=1}^{K_v} (\bvec{A}_k + \bvec{B}_k +\bvec{C}_k) \
	\bvec{F}_\textrm{in} \bvec{W}_k ,
\end{align}%eqnarray}
where $\bvec{F}_\textrm{in} \in \mathbb{R}^{N_v \times C_\textrm{in}}$ and $\bvec{F}_\textrm{out} \in \mathbb{R}^{N_v \times C_\textrm{out}}$ denote the input and output features, respectively.
%$N_v$ is the total number of human and object keypoints, i.e., $N_h+N_o$.
$C_\textrm{in}$ and $C_\textrm{out}$ are the number of input and output channels.
$\bvec{W}_k \in \mathbb{R}^{C_\textrm{in} \times C_\textrm{out}}$ denotes the weight matrix trainable through graph convolution and $K_v$ denotes the kernel size of the spatial dimension which is set to 3 by the spatial partitioning strategy.
$\bvec{A}_k$, $\bvec{B}_k$, and $\bvec{C}_k$ are the $N_v \times N_v$ matrices.
$\bvec{A}_k$ is a normalized adjacency matrix representing the three types of connections, which is calculated as
%physical connections in the human body, which is calculated as
%
$\bvec{A}_k = \bvec{\Lambda}_k^{-\frac{1}{2}} \overline{\bvec{A}}_k \bvec{\Lambda}_k^{-\frac{1}{2}}$ using the adjacency matrix $\overline{\bvec{A}}_k$.
$\overline{A}_k^{ij}$, the element $(i,j)$ of $\overline{\bvec{A}}_k$, is set to 1 or 0 depending on whether a keypoint $v_j$ is contained in the subset of keypoint $v_i$.
The diagonal elements of the normalized diagonal matrix $\bvec{\Lambda}_k$ are set as $\Lambda_k^{ii} = \sum_j(\overline{A}_k^{ij}) + \beta$, where $\beta$ is a small parameter to avoid empty rows.
$\bvec{B}_k$ is a trainable adjacency matrix that also represents the strength of the connections.
$\bvec{C}_k$ is a data-driven matrix based on the similarity of two keypoints.
%
%The input feature map $\bvec{F}_\textrm{in} \in \mathbb{R}^{C_\textrm{in} \times V}$, where $C_\textrm{in}$ and $V$ denote the number of input channels and keypoints, is embedded by two embedding functions $\zeta_k$ and $\eta_k$.
To calculate $\bvec{C}_k$, $\bvec{F}_\textrm{in}$ is first embedded by two embedding functions $\zeta_k$ and $\eta_k$, which are implemented as a $1 \times 1$ convolutional layer.
%
%Each function is implemented as an $1 \times 1$ convolutional layer.
%The output sizes of the functions are $(V,C')$ and $(C',V)$, and then they are multiplied to obtain a $V \times V$ similarity matrix.
%$\bvec{C}_k$ is obtained after a softmax normalization.
The series of calculations to obtain $\bvec{C}_k$ is written as $\bvec{C}_k = \textrm{softmax}(\bvec{F}_\textrm{in} \bvec{W}_{\zeta_k} \bvec{W}_{\eta_k}^\top \bvec{F}_\textrm{in}^\top)$, where $\bvec{W}_{\zeta_k}$ and $\bvec{W}_{\eta_k}$ denote the weight of the above embedding functions.
The utilization of three matrices, i.e., $\bvec{A}_k$, $\bvec{B}_k$, and $\bvec{C}_k$, makes it possible to lower the dependency on the predefined graphs.

An attention module called SKA is introduced, inspired by ST-JointAtt~\cite{song2021constructing}.
ST-JointAtt was inspired by~\cite{hou2021coordinate} and designed to distinguish the most informative keypoints in certain frames from the whole sequence by concurrently taking spatial and temporal information into account.
SKA is the attention module for capturing spatially informative keypoints.
The formulation of SKA is written as 
\begin{align}%eqnarray}
\label{e:att}
	\begin{aligned}
		\bvec{g}_\textrm{out} = \
		\bvec{g}_\textrm{in} \odot (\sigma(\theta (p(\bvec{g}_\textrm{in}) \cdot \bvec{W}_a) \cdot \bvec{W}_b)),
	\end{aligned}
\end{align}
where $\bvec{g}_\textrm{in}$ and $\bvec{g}_\textrm{out}$ denote input and output feature maps, which have the same channel size $C_g$.
$p(\cdot)$ denotes the average pooling operation.
$\theta(\cdot)$ and $\sigma(\cdot)$ denote the Hardswish and Sigmoid activation function, respectively.
$\bvec{W}_a \in \mathbb{R}^{C_g \times \frac{C_g}{4}}$ and $\bvec{W}_b \in \mathbb{R}^{\frac{C_g}{4} \times C_g}$ are trainable weight matrices.
$\odot$ denote the channel-wise outer-product.
This mechanism makes it possible to calculate attention maps to obtain informative human and object keypoints to be focus on.

%#####################################################################################################

\vspace{-2mm}
\section{EXPERIMENTS}
\label{s:exp}
%\vspace{-2mm}

\begin{figure*}[t!]
\centering
	%\vspace{-1mm}
	\includegraphics[width=1.0\linewidth]{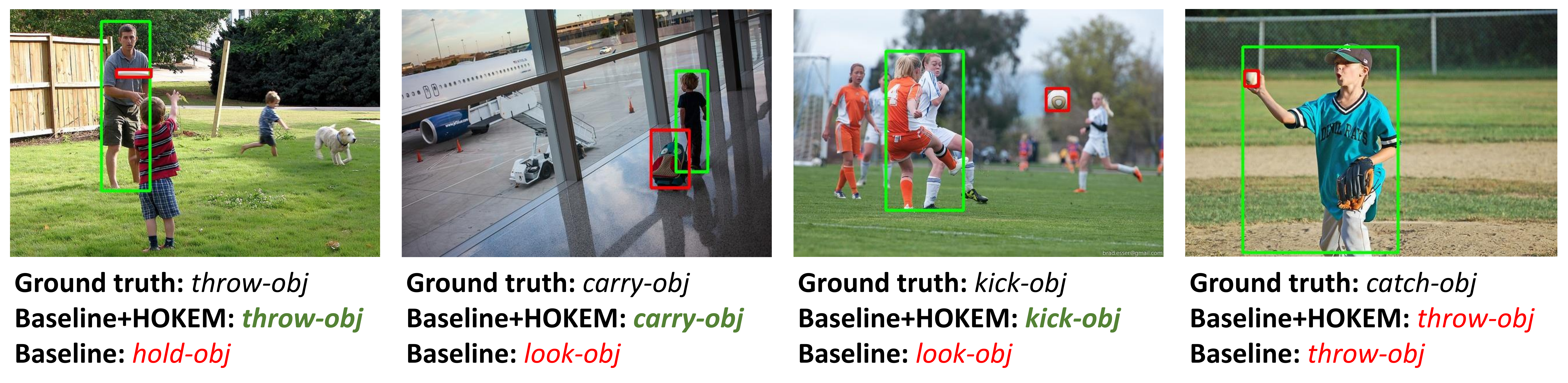}
	\vspace{-6mm}
	\caption{HOI detection results demonstrating the effectiveness of HOKEM. The second and third lines of the description are the HOI classes predicted by \textit{Baseline+HOKEM} and \textit{Baseline}, respectively. The three images on the left are examples that were successfully corrected to the ground truth using HOKEM, while the image on the right is an example that was not corrected.}
	\label{f:qual}
	\vspace{-4mm}
\end{figure*}

%\begin{comment}
\begin{table}[t]
	\begin{center}
	%\vspace{1mm}
	\caption{Comparisons of mAP with state-of-the-art methods. All methods utilize appearance information, and some also utilize pose information. Results that were not provided are marked with ``-''.}
	\label{t:sota}
	\vspace{-2mm}
	\scalebox{0.9}[0.85]{
  	\begin{tabular}{l c c c c}
  		\bhline{1.3pt}
	    Methods                                  & Backbone   & Pose & Sce. 1 & Sce. 2 \\ \hline
	    InteractNet~\cite{gkioxari2018detecting} & ResNet-50  & & 40.0 & 48.0 \\
	    iCAN~\cite{gao2018ican}                  & ResNet-50  & & 45.3 & 52.4 \\
	    UnionDet~\cite{kim2020uniondet}          & ResNet-50  & & 47.5 & 56.2 \\
	    VSGNet~\cite{ulutan2020vsgnet}           & ResNet-152 & & 51.8 & 57.0 \\ \hline
	    HOI Transformer~\cite{zou2021end}        & ResNet-101 & & 52.9 & - \\
	    HOTR~\cite{kim2021hotr}                  & ResNet-50  & & 55.2 & 64.4 \\ \hline
	    RPNN~\cite{zhou2019relation}             & ResNet-50  & \checkmark & -    & 47.5 \\
	    TIN~\cite{li2019transferable}            & ResNet-50  & \checkmark & 48.7 & - \\
	    PastaNet~\cite{li2020pastanet}           & ResNet-50  & \checkmark & 51.0 & 57.5 \\
	    PMFNet~\cite{wan2019pose}                & ResNet-50  & \checkmark & 52.0 & - \\
	    PD-Net~\cite{zhong2020polysemy}          & ResNet-152 & \checkmark & 52.0 & - \\
	    Liu et al.~\cite{liu2022human}           & ResNet-50-FPN & \checkmark & 52.3 & - \\
	    ACP~\cite{kim2020detecting}              & ResNet-152 & \checkmark & 53.0 & - \\
	    %FCMNet~\cite{liu2020amplifying}          & ResNet-50  & \checkmark & 53.1 & - \\
	    %SIGN~\cite{zheng2020skeleton}            & ResNet-50-FPN & \checkmark  & 53.1 & - \\
	    SGCN4HOI~\cite{zhu2022skeleton}          & ResNet-152 & \checkmark  & 53.1 & 57.9 \\ \hline
	    Baseline                                 & ResNet-152 & & 49.7 & 54.7 \\
	    Baseline + HOKEM                         & ResNet-152 & \checkmark  & 54.6 & 59.7 \\
	    \bhline{1.3pt}
    \end{tabular}
    }
    \end{center}
    \vspace{-8mm}
\end{table}

\vspace{-3mm}
\subsection{Dataset}
\label{ss:datasets}
\vspace{-2mm}

To evaluate the effectiveness of the proposed method, this paper adopted a common HOI benchmark, V-COCO~\cite{gupta2015visual}, which is a dataset created for HOI detection as a subset of MS-COCO~\cite{lin2014microsoft}.
V-COCO contains 5,400 images for train+val and 4,946 images for test.
The dataset includes 16,199 human instances in total and 24 HOI classes.
There are two evaluation methods in V-COCO, Scenarios 1 and 2.
If an object is occluded in an image, the bounding box is treated as empty in Scenario 1 and ignored in Scenario 2.

As with most conventional methods, average precision (AP) was used as an evaluation metric.
A HOI triplet $<$\textit{human-relation-object}$>$ is considered a true positive when both bounding boxes of the human and object have an IoU greater than 0.5 with the ground-truth boxes and the prediction label is correct.

\vspace{-3mm}
\subsection{Implementation Details}
\label{ss:imple}
\vspace{-2mm}

HO-AGCN was trained for 80 epochs in total.
A warmup strategy to gradually increase the learning rate from 0.0 to 0.1 was applied over the first 10 epochs for stable training.
After the 10\ts{th} epoch, the rate decreased to 0.0 by the 80\ts{th} epoch.
The stochastic gradient descent was applied as the optimizer and the momentum was set to 0.9.
The binary cross-entropy was selected as the loss function and the weight decay was set to 0.0001.
The batch size was set to 16.

HOKEM is an extension module for conventional HOI detection models.
To evaluate the performance of HOKEM, VSGNet~\cite{ulutan2020vsgnet} was selected as a representative conventional model.
Of the three branches utilized in VSGNet, the visual and spatial branches were utilized as the baseline model in the experiments, following~\cite{zhu2022skeleton}.
ResNet-152~\cite{he2016deep} was selected as the backbone feature extractor.
Seventeen human keypoints were extracted by DEKR~\cite{geng2021bottom}.
Mask R-CNN~\cite{he2017mask} was utilized for instance segmentation.

\begin{table}[t]
	\begin{center}
	\vspace{1mm}
	\caption{Performance comparison of networks using human and object keypoints. The proposed object keypoints, human-object graph, and keypoint-based features were utilized for all methods.}
	\label{t:comp_gcn}
	\vspace{-2mm}
	\scalebox{0.9}[0.85]{
  	\begin{tabular}{l c c c}
  		\bhline{1.3pt}
	    Method & Network & Sce. 1 & Sce. 2 \\ \hline
	    Baseline                           & - & 49.68 & 54.72 \\ \hline
	    Baseline + MLP                     & MLP & 50.20 & 55.33 \\
	    Baseline + ~\cite{liang2020pose}   & GCN & 50.60 & 55.54 \\ % PMN
	    Baseline + ~\cite{zhu2022skeleton} & GCN & 50.79 & 55.84 \\ \hline % Zhu
	    Baseline + HO-AGCN (w/o SKA)       & GCN & 54.03 & 59.22 \\
	    Baseline + HO-AGCN                 & GCN & \bf{54.57} & \bf{59.73} \\
	    \bhline{1.3pt}
    \end{tabular}
    }
    \end{center}
    \vspace{-5mm}
\end{table}

\begin{table}[t]
	\begin{center}
	%\vspace{1mm}
	\caption{Performance comparison of object keypoint extraction methods. HO-AGCN was utilized for all methods. \#keypoints denotes the number of object keypoints.} % The number of human keypoints is 17.
	\label{t:comp_okp}
	\vspace{-2mm}
	\scalebox{0.87}[0.85]{
  	\begin{tabular}{l c c c}
  		\bhline{1.3pt}
	    Object keypoint extraction method & \#keypoints & Sce. 1 & Sce. 2 \\ \hline
	    Baseline                              & - & 49.68 & 54.72 \\ \hline
	    Baseline + Vertex of bounding box     & 4 & 53.79 & 59.13 \\
	    Baseline + Center of bounding box~\cite{liang2020pose}  & 1 & 53.88 & 59.18 \\
	    Baseline + ~\cite{liu2022human}       & 9 & 54.00 & 59.16 \\ % Liu
	    Baseline + ~\cite{zhu2022skeleton}    & 9 & 54.39 & 59.30 \\ \hline % Zhu
	    Baseline + Proposed extraction method & 9 & \bf{54.57} & \bf{59.73} \\
	    \bhline{1.3pt}
    \end{tabular}
    }
    \end{center}
    \vspace{-8mm}
\end{table}

\vspace{-3mm}
\subsection{Quantitative Results and Ablation Studies}
\label{ss:results}
\vspace{-2mm}

The performance of HOKEM was evaluated when it was utilized with \textit{Baseline}.
Table~\ref{t:sota} shows the results and comparison with state-of-the-art methods in terms of mean average precision (mAP) score.
Compared with \textit{Baseline}, HOKEM significantly boosted the accuracy by approximately 5.0 mAP in both Scenarios.
The results obtained using HOKEM are as accurate as those of Transformer-based methods~\cite{zou2021end,kim2021hotr}, which have shown an improvement in accuracy over the past few years.
Moreover, in addition to higher accuracy than the conventional methods using human poses, another advantage of HOKEM is that it can be utilized in conjunction with many other methods as it simply multiplies the probabilities produced by the conventional methods.

Ablation studies were also carried out as shown in Tables~\ref{t:comp_gcn} and~\ref{t:comp_okp}.
Table~\ref{t:comp_gcn} shows comparisons of mAP between HO-AGCN and the conventional HOI detection networks using human and object keypoints.
To determine the effectiveness of HO-AGCN, the proposed object keypoints, human-object graph, and keypoint-based features were utilized in all of the methods except \textit{Baseline}+\cite{liang2020pose}, which only utilized center of box as the object keypoint.
MLP consists of five layers with 256 nodes in the hidden layers.
The MLP and GCNs proposed for HOI~\cite{liang2020pose,zhu2022skeleton} slightly increased the accuracy of \textit{Baseline} by approximately 1.0 mAP, whereas the proposed method improved the accuracy by approximately 4.5 mAP.
The results show the effectiveness of the structure of HO-AGCN and the AGC layer reducing the topology dependence for the predefined graph.
The SKA layer also further improved the accuracy by more than 0.5 mAP, demonstrating the effectiveness of considering spatially informative keypoints using the attention mechanism.

Table~\ref{t:comp_okp} compares the mAP of methods using object keypoints. % are shown.
To determine the effectiveness of the proposed object keypoint extraction method, HO-AGCN was utilized in all of the methods.
As shown in the table, the proposed extraction method achieved the highest accuracy compared to when using the vertex or center of the bounding box~\cite{liang2020pose} and the conventional methods~\cite{liu2022human,zhu2022skeleton}.
This result indicates that the proposed extraction method, which represents the gravity and contour of objects on the basis of a segmentation mask, is robust for a wide variety of object shapes.

\vspace{-3mm}
\subsection{Qualitative Results}
\label{ss:results}
\vspace{-2mm}

The AP of 24 HOI classes in Scenario 1 were also compared between \textit{Baseline} and \textit{Baseline+HOKEM}, and HOKEM improved the AP in all classes.
The top three interactions with the large AP gains were \textit{throw-obj} (+9.4), \textit{carry-obj} (+8.7), and \textit{kick-obj} (+8.0).
Figure~\ref{f:qual} shows the qualitative results which verify the effectiveness of HOKEM.
In the leftmost image, it is presumed that the object is detected as being \textit{thrown} rather than \textit{held} because HOKEM explicitly calculated the distance between the keypoints of the hand and the object as an input feature.
Similarly, the corrections are presumed to be successful because the second image from the left captured the front-back positioning between the person and the object, and the third image from the left captured the kicking pose and the position between the person and the object.
However, the ground truth of the rightmost image is \textit{catching} an object, and even though the proposed method was utilized, the person was still estimated to be \textit{throwing} an object as with \textit{Baseline}.
This example implies that capturing the temporal changes of the person and the object is important for more practical HOI detection.
Therefore, it can be concluded that HOKEM, which introduces the proposed object keypoint extraction method and HO-AGCN, is effective in increasing the accuracy of the conventional appearance-based model.

%#####################################################################################################

%\vspace{-1mm}
\section{CONCLUSION}
\label{s:concl}
\vspace{-1mm}

This paper presented HOKEM, an easy-to-use extension module to boost the accuracy of appearance-based HOI detection models.
HOKEM consists of a novel object keypoint extraction method and a graph convolutional network, HO-AGCN.
The proposed object keypoint extraction method represents the contour of objects in a robust manner for a wide variety of object shapes using a segmentation mask.
HO-AGCN accurately captures the spatial structure of the human and object through the topology optimization of the predefined human-object graph and the attention mechanism.
Through the experiments of HOI detection, the proposed method showed a significant improvement in accuracy over the conventional methods.

% using a HOI detection dataset

%#####################################################################################################

\vfill
\pagebreak

\end{document}